\documentclass[final]{cvpr}

\usepackage{times}
\usepackage{graphicx}
\usepackage{grffile}
\usepackage{amsmath}
\usepackage{amssymb}
\usepackage{multirow}
\usepackage[pagebackref=true,breaklinks=true,colorlinks,bookmarks=false]{hyperref}
\usepackage[export]{adjustbox}
\usepackage{subcaption}
\usepackage{tikz}
\usepackage{pgfplots}

\usepackage[inline]{enumitem}
\usepackage{booktabs}
\usepackage[pagebackref=true,breaklinks=true,colorlinks,bookmarks=false]{hyperref}

\newcommand{\bs}{\boldsymbol}

\newcommand{\Xa}{\bs{X}_{1}}
\newcommand{\Xb}{\bs{X}_{2}}

\definecolor{mygray}{HTML}{C6C6C6}
\definecolor{myblue}{HTML}{1A6EAA}
\definecolor{mypurple}{HTML}{6f00CC}

\newcommand{\hl}[1]{{\color{mypurple}#1}}

\newcommand{\comment}[1]{{\color{orange}#1}}

\newcommand{\remove}[1]{}

\begin{document}

\title{NBNet: Noise Basis Learning for Image Denoising with Subspace Projection}

\author{
    Shen Cheng$^{1}$\ \ \ Yuzhi Wang$^{1}$\ \ \ Haibin Huang$^{2}$\ \ \ Donghao Liu$^{1}$\ \ \ Haoqiang Fan$^{1}$\ \ \ Shuaicheng Liu$^{3,1}$\thanks {Corresponding author}
	\\
	\\
    $^{1}$Megvii Technology \quad  $^{2}$Kuaishou Technology \\
    $^{3}$University of Electronic Science and Technology of China \\
    \url{https://github.com/megvii-research/NBNet}
}


\maketitle
\thispagestyle{empty}
\begin{abstract}
   In this paper, we introduce NBNet, a novel framework for image denoising. Unlike previous works, we propose to tackle this challenging problem from a new perspective: noise reduction by image-adaptive projection. Specifically, we propose to train a network that can separate signal and noise by learning a set of reconstruction basis in the feature space. Subsequently, image denosing can be achieved by selecting corresponding basis of the signal subspace and projecting the input into such space. Our key insight is that projection can naturally maintain the local structure of input signal, especially for areas with low light or weak textures.  Towards this end,  we propose 
   SSA, a non-local attention module we design to explicitly learn the basis generation as well as subspace projection. We further incorporate SSA with NBNet, a UNet structured network designed for end-to-end image denosing based. We conduct evaluations on benchmarks, including SIDD and DND, and NBNet achieves state-of-the-art performance on PSNR and SSIM with significantly less computational cost.  
\end{abstract}

\section{Introduction}
\begin{figure}[t]
    \centering
    \includegraphics[width=1.0\linewidth]{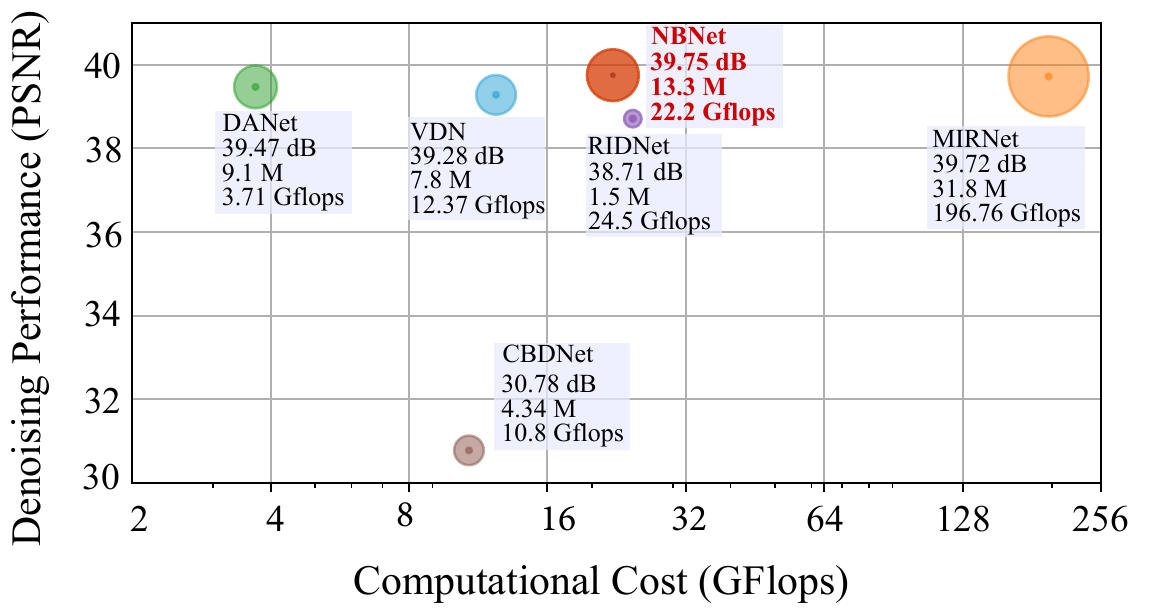}
    \caption{PSNRs at different computational cost and parameter amount of our method and previous methods in SIDD ~\cite{abdelhamed2018high}. The proposed NBNet achieves SOTA performance with a balanced computational requirement.}
    \label{fig:flop_psnr}
\end{figure}

\begin{figure}[t]
    \centering
    \includegraphics[width=1.0\linewidth]{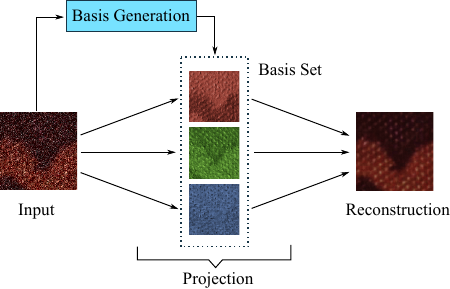}
    \caption{Denoising via subspace projection: Our NBNet learns to generate a set of basis for the signal subspace and by projecting the input into this space, signal can be enhanced after reconstruction for easy separation from noise.}
    \label{fig:projection_concept}
\end{figure}
Image denoising is a fundamental and long lasting task in image processing and computer vision. The main challenging is to recover a clean signal $\bs{x}$ from the noisy observation $\bs{y}$, with the additive noise $\bs{n}$, namely:
\begin{equation}
    \bs{y} = \bs{x} + \bs{n}
\end{equation}
This problem is ill-posed as both the image term $\bs{x}$ and the noise term $\bs{n}$ are unknown and can hardly be separated. Towards this end, many denoising methods utilize image prior and a noise model to estimate either image or noise from the noisy observation. For example, traditional methods such as NLM~\cite{buades2005non} and BM3D~\cite{dabov2007image} use the local similarity of image and the independence of noise, and wavelet denoising~\cite{portilla2003image} utilizes the sparsity of image in transformed domain. 

Recent deep neural networks~(DNN) based denoising methods~\cite{tai2017memnet,chen2017trainable,zhou2019awgn,jain2009natural,xie2012image,mao2016image,ulyanov2018deep} usually implicitly utilize image prior and noise distribution learned from a large set of paired training data. \remove{Although previous DNNs based methods have achieved tremendous success, it still remains challenging to restore high quality images from extremely noisy ones where the noise is hard to model. Moreover, it is also problematic for those methods to distinguish flatten regions and area with weak textures, learning to over-smooth or insufficient denoising. }

\begin{figure*}[h]
    \begin{center}
        \includegraphics[width=.9\linewidth]{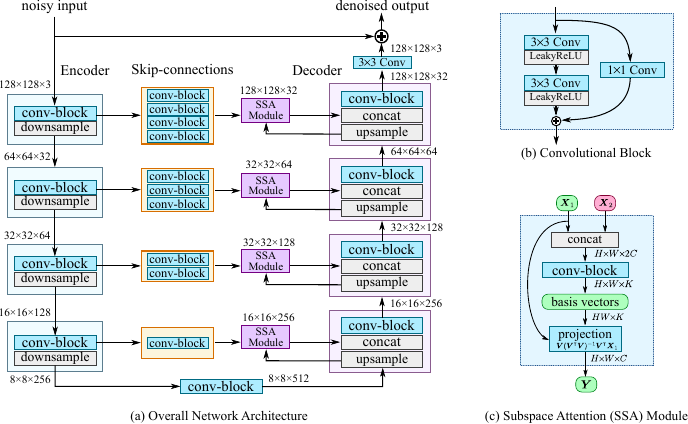}
    \end{center}
    \caption{Overall architecture of NBNet and structure of key building blocks. NBNet is based on UNet architecture with a depth of 5 and our SSA module is used to project features of skip-connection from the encoder. }
    \label{fig:architecture}
\end{figure*}

Although previous CNN-based methods have achieved tremendous success, it is still challenging to recover high quality images in hard scenes such as weak textures or high-frequency details. Our key observation is that convolutional networks usually depend on local filter response to separate noise and signal. While in hard scenes with low signal-noise-ratio~(SNR), local response can easily get confused without additional global structure information. 

In this paper, we utilize non-local image information by \emph{projection}. The basic concept of image projection is illustrated in Fig.~\ref{fig:projection_concept}, where a set of image basis vectors are generated from the input image, then we reconstruct the image inside the subspace spanned by these basis vectors. As natural images usually lie in a low-rank \emph{signal subspace}, by properly learning and generating the basis vectors, the reconstructed image can keep most original information and suppress noise which is irrelevant to the generated basis set. 
Based on this idea, we propose NBNet, depicted in Fig.~\ref{fig:architecture}. The overall architecture of NBNet is a commonly-used UNet~\cite{ronneberger2015u}, except for the crucial ingredient \textbf{s}ub\textbf{s}pace \textbf{a}ttention~(SSA) module which learns the subspace basis and image projection in an end-to-end fashion. Our experiments on popular benchmark datasets such as SIDD~\cite{abdelhamed2018high} and DnD~\cite{plotz2017benchmarking} demonstrate that the proposed SSA module brings a significant performance boost in both PSNR and SSIM with much smaller computational cost than adding convolutional blocks. As depicted in Fig.~\ref{fig:flop_psnr}, the whole architecture of NBNet achieves the state-of-the-art performance while only a smaller additional computational cost is added. To summarize, our contributions include:
\begin{itemize}
\item We analyze the image denoising problem from a new perspective of subspace projection. We further design a simple and efficient SSA module to learn subspace projection which can be plugged into normal CNNs.
\item We propose NBNet, a UNet with SSA module for projection based image denoising. 

\item NBNet acheives state-of-the-art performance in PSNR and SSIM on many popular benchmarks
\item We provide in-depth analysis of projection based image denoising,  demonstrating it is a promising direction to explore. 
\end{itemize}

\section{Related Works}

\begin{figure*}[h]
\centering
\scalebox{1.02}{\hspace{-2.6mm}
   \begin{tabular}[t]{c@{ }c@{ }c@{ }c@{ }c@{ }c@{ }c@{ }}
   \includegraphics[width=.16\textwidth]{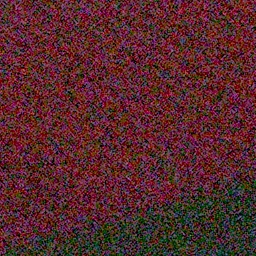}& \hspace{-1.4mm}
   \includegraphics[width=.16\textwidth]{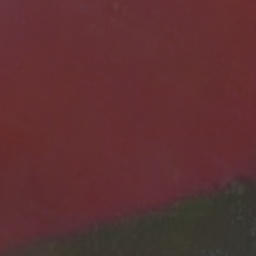}&  \hspace{-1.4mm}
   \includegraphics[width=.16\textwidth]{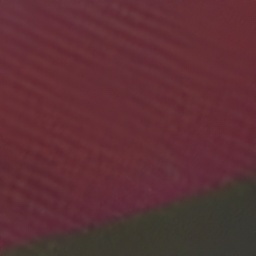}& \hspace{-1.4mm}
   \includegraphics[width=.16\textwidth]{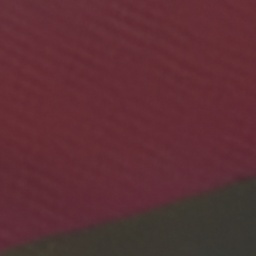}& \hspace{-1.4mm}
   \includegraphics[width=.16\textwidth]{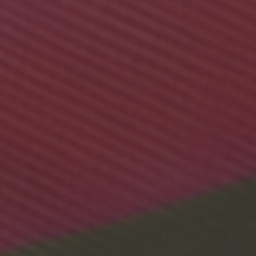}&  \hspace{-1.4mm}
   \includegraphics[width=.16\textwidth]{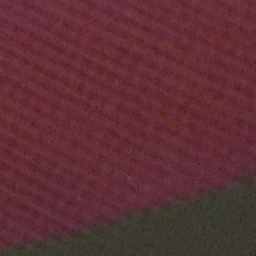}
   \vspace{0.2mm}
   \\
   \vspace{0.5mm}
    19.11 dB & 33.63 dB & 34.27 dB & 34.29 dB & \textbf{34.70 dB} &\\
   \includegraphics[width=.16\textwidth]{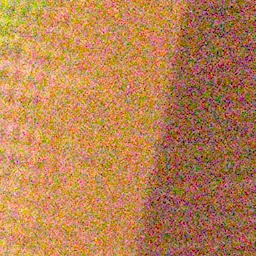}& \hspace{-1.4mm}
   \includegraphics[width=.16\textwidth]{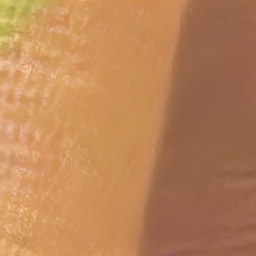}&  \hspace{-1.4mm}
   \includegraphics[width=.16\textwidth]{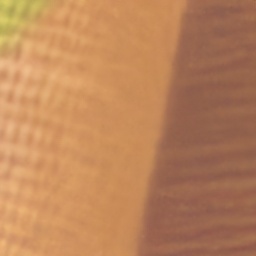}& \hspace{-1.4mm}
   \includegraphics[width=.16\textwidth]{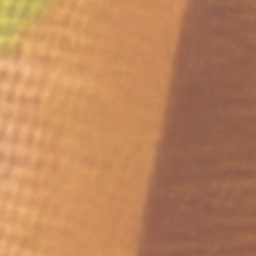}& \hspace{-1.4mm}
   \includegraphics[width=.16\textwidth]{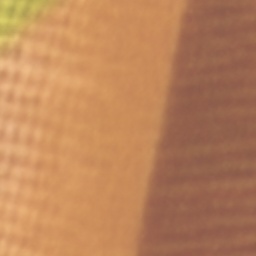}&  \hspace{-1.4mm}
    \includegraphics[width=.16\textwidth]{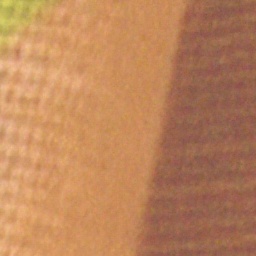}
   \vspace{0.2mm}
   \\
   \vspace{0.5mm}
    19.48 dB & 32.62 dB & 34.14 dB & 33.94 dB & \textbf{34.40 dB} &\\
   \includegraphics[width=.16\textwidth]{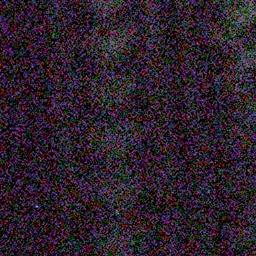}& \hspace{-1.4mm}
   \includegraphics[width=.16\textwidth]{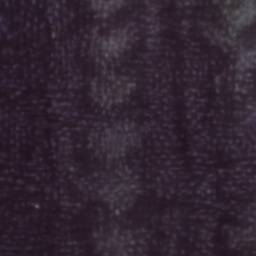}&  \hspace{-1.4mm}
   \includegraphics[width=.16\textwidth]{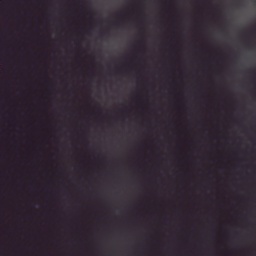}& \hspace{-1.4mm}
   \includegraphics[width=.16\textwidth]{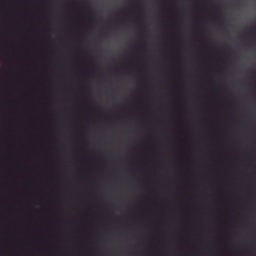}& \hspace{-1.4mm}
   \includegraphics[width=.16\textwidth]{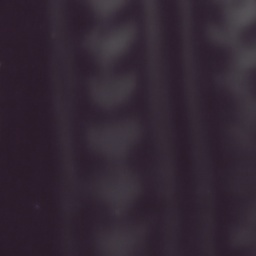}&  \hspace{-1.4mm}
    \includegraphics[width=.16\textwidth]{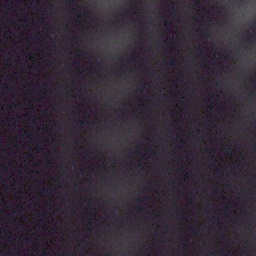}
   \vspace{0.2mm}
   \\
   \vspace{0.5mm}
    19.49 dB & 30.06 dB & 32.17 dB & 32.00 dB & \textbf{32.68 dB} &\\
    Noisy
   & VDN~\cite{yue2019variational}  & DANet~\cite{yue2020dual} & MIRNet~\cite{zamir2020learning} & Ours & Reference \\
   \end{tabular}
}
   \vspace{-2mm}
   \caption{\small Denoising examples from SIDD~\cite{abdelhamed2018high}. Our results superior in weak texture areas like dots and lines pattern}
   \label{fig:sidd example}
   
\end{figure*}

 \begin{table*}[t]
    \begin{center}
    \setlength{\tabcolsep}{2.5pt}
    \scalebox{0.90}{
    \begin{tabular}{l c c c c c c c c c c c c c c c}
    \toprule
    Method & DnCNN & MLP     & FoE   & BM3D    & WNNM    & NLM    & KSVD    & EPLL  & CBDNet & RIDNet &  VDN &
     DANet & MIRNet & NBNet\\
     & ~\cite{zhang2017beyond} & ~\cite{burger2012image} & ~\cite{rudin1992nonlinear} & ~\cite{dabov2006image} & ~\cite{gu2014weighted} & ~\cite{buades2005non} & ~\cite{aharon2006k} &~\cite{zoran2011learning} & ~\cite{guo2018toward} & ~\cite{anwar2019ridnet} & ~\cite{yue2019variational} &  ~\cite{yue2020dual} & ~\cite{zamir2020learning} & ours\\
    \midrule
    PSNR~$\textcolor{black}{\uparrow}$ &  23.66  &   24.71  &    25.58  & 25.65  &   25.78  &   26.76  &  26.88  & 27.11  &  30.78  & 38.71 & 39.28 & 39.47 &
    39.72 & \textbf{39.75}\\
    SSIM~$\textcolor{black}{\uparrow}$ &  0.583 &  0.641 &    0.792 &  0.685 &  0.809 &  0.699 &  0.842 &  0.870 & 0.754   &    0.914 &  0.909  &  0.918 &
    0.959 & \textbf{0.973}\\
    \bottomrule
    \end{tabular}}
    \end{center}
    \vspace{-6mm}
    \caption{\small Denoising comparisons on the SIDD~\cite{abdelhamed2018high} dataset .}
    \label{table:sidd}
\end{table*}
\begin{figure*}[h]
\begin{center}
\scalebox{1.0}{
\begin{tabular}[b]{c@{ } c@{ }  c@{ } c@{ } c@{ }	}\hspace{-3.7mm}
     \multirow{4}{*}{\includegraphics[width=.364\textwidth,valign=tl]{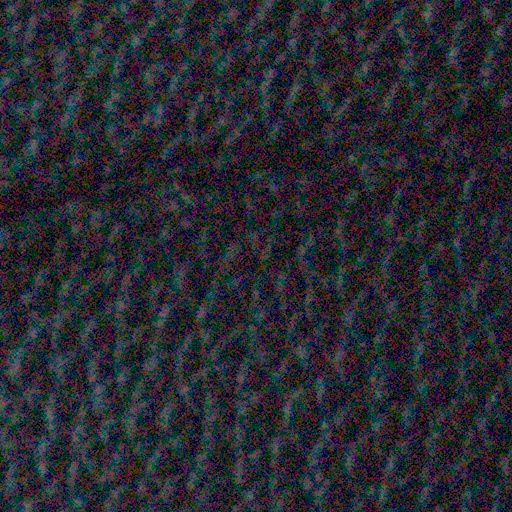}} 
    &  \includegraphics[trim={10.8cm 6.5cm  3cm  7.2cm },clip,width=.152\textwidth,valign=t]{pic/dnd_results/Noisy.jpg}
    & \includegraphics[trim={10.8cm 6.5cm  3cm  7.2cm },clip,width=.152\textwidth,valign=t]{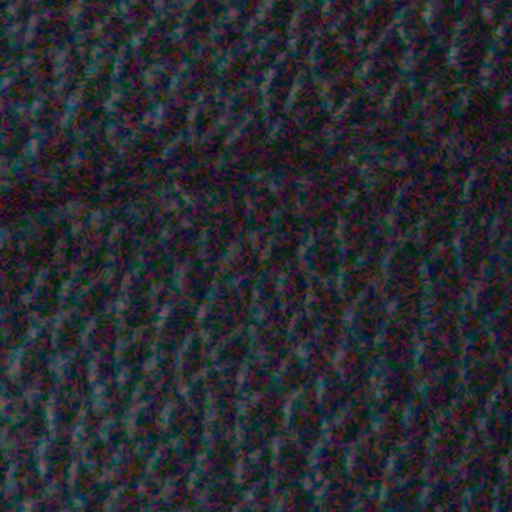}
    & \includegraphics[trim={10.8cm 6.5cm  3cm  7.2cm },clip,width=.152\textwidth,valign=t]{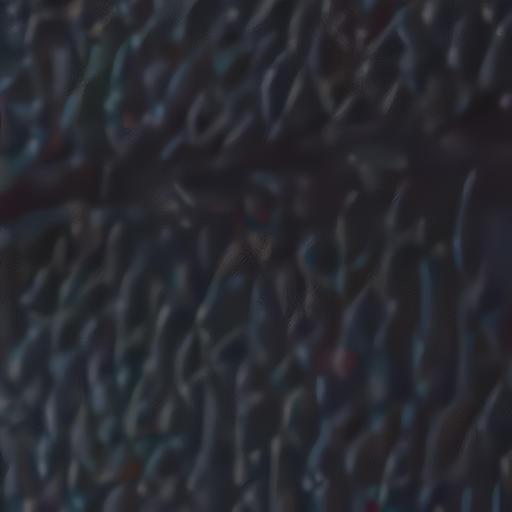}
    & \includegraphics[trim={10.8cm 6.5cm  3cm  7.2cm },clip,width=.152\textwidth,valign=t]{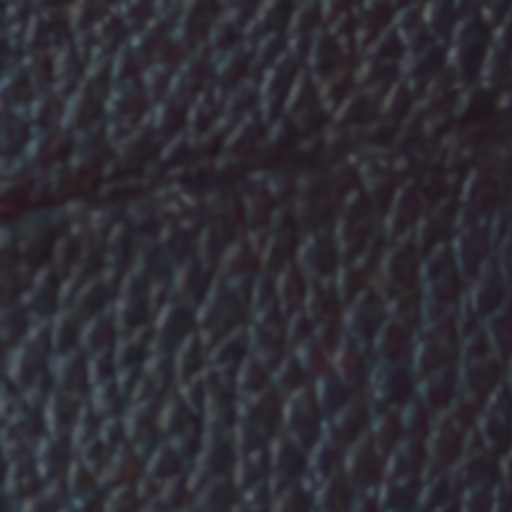}

    \vspace{0.5mm}
\\
    &      &25.65 dB  & 31.54 dB
    & 32.91 dB\\
    & Noisy &BM3D~\cite{dabov2007image} &  FFDNet~\cite{zhang2018ffdnet} & DANet~\cite{yue2020dual}
    \\
    &\includegraphics[trim={10.8cm 6.5cm  3cm  7.2cm },clip,width=.152\textwidth,valign=t]{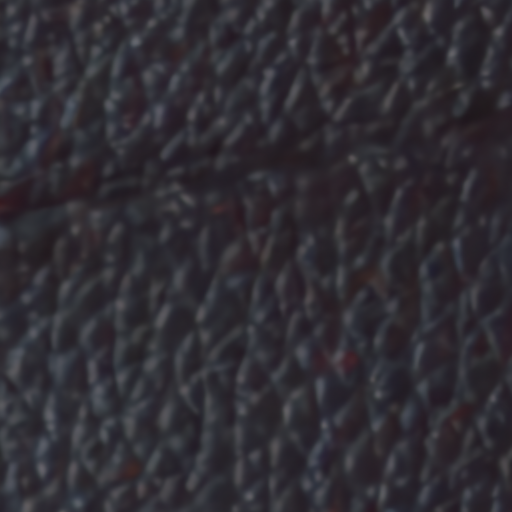}  
    &\includegraphics[trim={10.8cm 6.5cm  3cm  7.2cm },clip,width=.152\textwidth,valign=t]{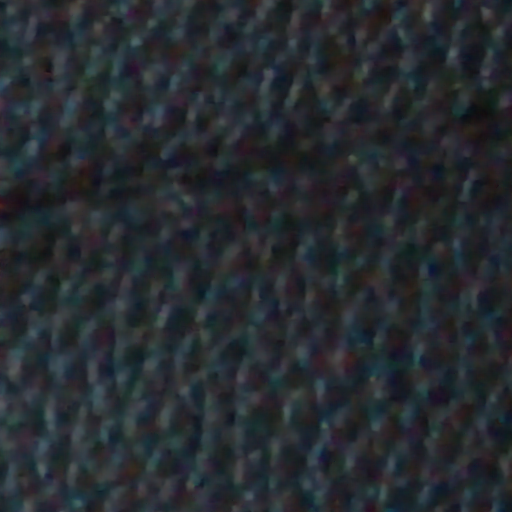}&
     \includegraphics[trim={10.8cm 6.5cm  3cm  7.2cm },clip,width=.152\textwidth,valign=t]{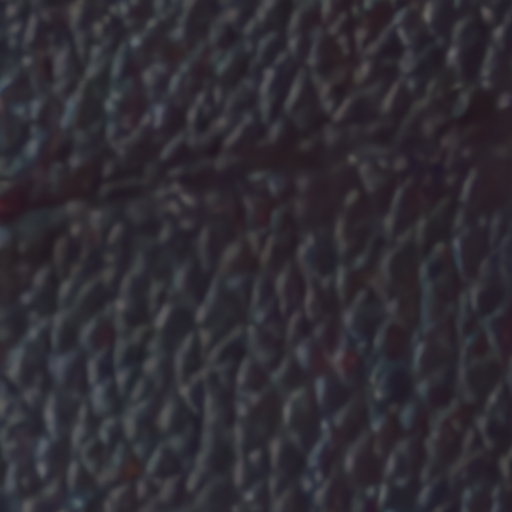}&
     \includegraphics[trim={10.8cm 6.5cm  3cm  7.2cm },clip,width=.152\textwidth,valign=t]{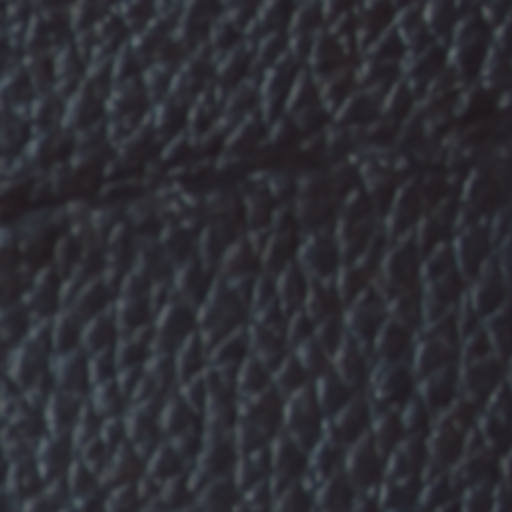}
     \vspace{0.5mm}
     \\

      & 32.55 dB 
     & 31.54 dB
     & 32.09 dB & \textbf{34.12 dB}\\
           Noisy Image  & MIRNet~\cite{zamir2020learning} & CBDNet~\cite{guo2018toward}  
           & VDN~\cite{yue2019variational}     & Ours \\
\end{tabular}}
\end{center}
\vspace{-6mm}
\caption{\small Denoising examples from DND~\cite{plotz2017benchmarking}. Our results preserve the textures and sharpness. }
\label{fig:dnd example}
\end{figure*}
\begin{table*}[t]
    \setlength{\tabcolsep}{2.5pt}
    \centering
    {
    \scalebox{0.90}{
    \begin{tabular}{l c c c c c c c c c c c c}
    \toprule
     Method & BM3D  & KSVD    & MCWNNM & FFDNet+ & TWSC  & CBDNet & RIDNet  & VDN & DANet &
     MIRNet & NBNet\\
     & ~\cite{dabov2006image} & ~\cite{aharon2006k} & ~\cite{xu2017multi} & ~\cite{zhang2018ffdnet} & ~\cite{xu2018trilateral} &~\cite{guo2018toward} &~\cite{anwar2019ridnet} & ~\cite{yue2019variational} & ~\cite{yue2020dual} & ~\cite{zamir2020learning} & ours\\
    \midrule
    PSNR~$\textcolor{black}{\uparrow}$ & 34.51 & 36.49 &  37.38 & 37.61 & 37.94 & 38.06 &  39.26 & 39.38 & 39.59
    & 39.88 & \textbf{39.89} \\
    SSIM~$\textcolor{black}{\uparrow}$& 0.851 & 0.898 & 0.929 & 0.942 & 0.940 & 0.942 & 0.953 & 0.952 & 0.955
    & \textbf{0.956} & \emph{0.955} \\
    \bottomrule
    \end{tabular}}}
    \vspace*{-2mm}\caption{\small Denoising comparisons on the DND~\cite{plotz2017benchmarking} dataset.}
     \label{table:dnd-result}
 \end{table*}

\subsection{Traditional Methods}

Image noise reduction is a fundamental component in image processing problem and has been studied for decades. Early works usually rely on image priors, including non-local means~(NLM) \cite{buades2005non}, sparse coding \cite{elad2006image,mairal2009non,aharon2006k}, 3D transform-domain filtering~(BM3D) \cite{dabov2007image}, and others \cite{gu2014weighted,portilla2003image}. Although these classical approaches like BM3D, can generate reasonable desnoising results with certain accuracy and robustness, their algorithmic complexity is usually high and with limited generalization. With the recent development of convolutional neural networks~(CNNs),  end-to-end trained denoising CNNs has gained considerable attention with great success in this field. 

\subsection{Network architecture}

One main stream of CNNs based desnoising is to design novel network architecture to tackle this problem. Earlier work~\cite{burger2012image} proposed to apply multi-layer perceptron~(MLP) to denoising task and achieved comparable results with BM3D. Since then more advanced network architectures are introduced. Chen et al.~\cite{chen2017trainable} proposed a trainable nonlinear reaction diffusion~(TNRD) model for Gaussian noise removal at different level. DnCNN~\cite{zhang2017beyond} demonstrated the effectiveness of residual learning and batch normalization for denoising network using deep CNNs.  Later on, More network structures were proposed to either enlarge the receptive field or balance the efficiency, like dilated convolution~\cite{zhang2017learning}, autoencoder with skip connection~\cite{mao2016image}, ResNet~\cite{ren2018dn}, recursively branched deconvolutional network~(RBDN)~\cite{santhanam2017generalized}. Recently some interests are put into involving high-level vision semantics like classification and segmentation with image denoising. Works~\cite{liu2017image, niknejad2017class} applied segmentation to enhance the denoising performance on different regions. \cite{zhang2018ffdnet} recently proposed FFDNet, a non-blind denoising by concatenating the noise level as a map to the noisy image and demonstrated a spatial-invariant denoising on realistic noises with over-smoothed detail. MIRNet~\cite{zamir2020learning} proposed a general network architecture for image enhancement such as denoising and super-resolution with many noval
build blocks which can extract, exchange and utilize multi-scale feature information.

In this work, we adapt a UNet style architecture with a novel subspace attention module. Unlike ~\cite{anwar2019ridnet, Anwar2020IERD,tian2020attention} use attention module for region or feature selection, SSA is designed to learn the subspace basis and image projection. 

%

\subsection{Noise distribution}

To train the deep networks mentioned above, it requires high quality real datasets with a huge amount of clean and noisy image pairs, which is hard and  tedious to construct in practice. Hence, the problem of synthesizing realistic image noise has also been extensively studied.  To approximate real noise, multiple types of synthetic noise are explored in previous work, such as Gaussian-Poisson~\cite{foi2008practical,liu2014practical}, in-camera process simulation~\cite{liu2008automatic,guo2018toward}, Gaussian Mixture Model~(GMM)~\cite{zhu2016noise} and GAN-generated noises~\cite{chen2018image} and so on. It has been shown that networks properly trained from the synthetic data can generalize well to real data~\cite{zhou2019awgn,brooks2019unprocessing,PDRID}. Different from all the aforementioned works that focus on noise modeling , our method study subspace basis generation and improve noise reduction by projection. 
\section{Method}
\begin{table*}[h]
    \begin{center}

    \setlength{\tabcolsep}{2.5pt}
    \begin{tabular}{c c c c c c c c c c c c c}
        \hline
        \multirow{3}*{Cases}&\multirow{3}*{Datasets} & \multicolumn{11}{c}{Methods} \\
        \cline{3-13}
        & & CBM3D & WNNM    & NCSR  &MLP  &DnCNN-B &MemNet  &FFDNet
        &$\text{FFDNet}_v$ &UDNet &VDN & Ours\\
        & & ~\cite{guo2018toward} & ~\cite{gu2014weighted}    & ~\cite{dong2013nonlocally}  &~\cite{burger2012image}  &~\cite{zhang2017beyond} &~\cite{tai2017memnet}  &~\cite{zhang2018ffdnet}
        &~\cite{zhang2018ffdnet} &~\cite{lefkimmiatis2018universal} &~\cite{yue2019variational} & \\
        \hline
        \multirow{3}*{Case 1} & Set5
                    & 27.76 &26.53   &26.62  &27.26  &29.85     & 30.10   &30.16  &30.15  &28.13  &\textit{30.39} & \textbf{30.59}\\
        \cline{2-13}                                            
          &  LIVE1  & 26.58 &25.27   &24.96  &25.71  &28.81     & 28.96   &28.99  &28.96  &27.19  &\textit{29.22} & \textbf{29.40} \\
        \cline{2-13}                                          
          &  BSD68  & 26.51 &25.13   &24.96  &25.58  &28.73     & 28.74   &28.78  &28.77  &27.13  &\textit{29.02} & \textbf{29.16}\\
        \hline                                                   
        \multirow{3}*{Case 2} & Set5                                     
                    & 26.34 &24.61   &25.76  &25.73  &29.04     & 29.55   &29.60  &29.56  &26.01  &\textit{29.80} & \textbf{29.88}\\
        \cline{2-13}                                            
          &  LIVE1  & 25.18 &23.52   &24.08  &24.31  &28.18     & 28.56   &28.58  &28.56  &25.25  &\textit{28.82} &\textbf{29.01} \\
        \cline{2-13}                                             
          &  BSD68  & 25.28 &23.52   &24.27  &24.30  &28.15     & 28.36   &28.43  &28.42  &25.13  &\textit{28.67}  & \textbf{28.76}\\
        \hline                                                 
        \multirow{3}*{Case 3} & Set5                                     
                    & 27.88 &26.07   &26.84  &26.88  &29.13     & 29.51   &29.54  &29.49  &27.54  &\textit{29.74} & \textbf{29.89}\\
        \cline{2-13}                                           
        &  LIVE1    & 26.50 &24.67   &24.96  &25.26  &28.17     & 28.37   &28.39  &28.38  &26.48  &\textit{28.65} & \textbf{28.82}\\
        \cline{2-13}                                             
        &  BSD68    & 26.44 &24.60   &24.95  &25.10  &28.11     & 28.20   &28.22  &28.20  &26.44  &\textit{28.46}  &  \textbf{28.59}\\
        \hline
    \end{tabular}
    \end{center}
    \vspace{-6mm}
    \caption{\small{The PSNR~(dB) results of all competing methods on the three groups of test datasets. The best and second
best results are highlighted in bold and Italic, respectively.}}
    \label{tab:psnr_noniid}

\end{table*}

\remove{
\hl{Denote a noisy image $y=x+n$, where $x$ is the underlying clean image~(the signal) and $n$ is the noise, by feeding it into a neural network of sufficient capacity, we can project y towards the intermediate feature maps. Since neural networks are highly nonlinear, it is easier to distinguish between signal and noise in deep feature space. In this way, we can estimate a set of basis vectors to reconstruct $x$ from $y$. Specifically, some vectors represent signal while others represent noise, so we can reduce noise and preserve signal. To this aim, we first propose the subspace attention~(SSA) module, which estimates such basis vectors in the deep feature space.}
\comment{needs re-writing.}
Concretely, we choose the orthogonal projection as the transformation because the basis vectors are not restricted to orthogonality. Similarly the non-local module takes the attention maps as $\bs{\Phi}$ and the dot production as transformation. In the case of non-orthogonal basis vectors, the feature maps might be broken and we will show by experiments that it doesn't apply to our method. However when the feature map size is large, the matrix $\bs{\Phi}$ will be a squared multiple of the feature map size, our proposed subspace attention module can avoid increasing parameters and reduce computational cost compared to non-local module.
}

\subsection{Subspace Projection with Neural Network}

As shown in Fig.~\ref{fig:projection_concept}, the projection contains two main steps:
\begin{enumerate}[label=\alph*),topsep=0pt, partopsep=0pt,itemsep=0pt,parsep=0pt]
	\item \emph{Basis generation}: generating subspace basis vectors from image feature maps;
	\item \emph{Projection}: transforming feature maps into the signal subspace.
\end{enumerate}

We denote $\Xa, \Xb \in \mathbb{R}^{H\times W \times C}$  as two feature maps from a single image. 
They are the intermediate activations of a CNN and can be in different layers but with the same size.
We first estimate $K$ basis vectors
$[\bs{v}_{1},\bs{v}_{2}, \cdots, \bs{v}_{K}]$
based on $\Xa$ and $\Xb$, and each $\bs{v}_{i}\in \mathbb{R}^{N}$ is a basis vector of the signal subspace, where $N=HW$. 
Then we transform $\Xa$ into the subspace spanned by $\{\bs{v}\}$.

\subsubsection{Basis Generation}
Let $f_{\theta}: (\mathbb{R}^{H\times W\times {C}}, \mathbb{R}^{H\times W\times {C}}) \rightarrow \mathbb{R}^{N\times K}$ be a function parameterized by $\theta$, basis generation can be written as:
\begin{equation}
	\bs{V}=f_{\theta}(\Xa, \Xb) \text{,}
	\label{eq: basis_func}
\end{equation}
where  $\Xa$ and $\Xb$ are image feature maps and $\bs{V} = [\bs{v}_{1},\bs{v}_{2}, \cdots, \bs{v}_{K}]$ is a matrix
composed of basis vectors. 
We implement the function $f_\theta$ with a small convolutional network. We first concatenate $\Xa$ and $\Xb$
along the channel axis as $\bs{X} \in \mathbb{R}^{H \times W \times 2C}$, then feed it into a shallow residual-convolutional block with $K$ output channels (Fig.~\ref{fig:architecture}(b)), whose output can then be reshaped to $HW\times K$. The weights and biases of the basis generation blocks are updated during the training in an end-to-end fashion.

 

\subsubsection{Projection}
Given the aforementioned matrix $\bs{V}\in\mathbb{R}^{N\times K}$ whose columns are basis vectors of a 
$K$-dimensional signal subspace $\mathcal{V} \subset \mathbb{R}^{N}$, 
we can project the image feature map $\Xa$ onto $\mathcal{V}$ by orthogonal linear projection. 

Let $\bs{P}: \mathbb{R}^{N}\rightarrow \mathcal{V}$ be the orthogonal projection matrix to signal subspace, $\bs{P}$ can calculated from $\bs{V}$~\cite{meyer2000}, given by
\begin{equation}
    \bs{P} = \bs{V}(\bs{V}^\intercal\bs{V})^{-1}\bs{V}^\intercal,
\end{equation}
where the normalization term $(\bs{V}^\intercal\bs{V})^{-1}$ is required since the basis generation
process does not ensure the basis vectors are orthogonal to each other.

Finally, the image feature map $\Xa$ can be reconstructed in the signal subspace by as $\bs{Y}$, given
by
\begin{equation}
    \bs{Y} = \bs{P}\Xa \text{.}
\end{equation}

The operations in projection are purely linear matrix manipulations with some proper reshaping, which
is fully differentiable and can be easily implemented in modern neural network frameworks.

Combining basis generation and subspace projection, we construct the structure of the proposed SSA module, illustrated in Fig.~\ref{fig:architecture}(c).

\remove{
The transformation step $T$ works as change of basis. In general, the transformed feature map can be represented in the following form:
\begin{equation}
    \bs{Y} = T(\bs{V}, \Xa)
\end{equation}
Here, we take the orthogonal projection as $T$. Note that the column vectors of $\bs{\Phi}$ are all zero except $\bs{\Phi}$, the component of $\Xa$ on the $\Vec{0}$ vectors does not have any subsequent effect on the network, so we can use $\bs{U}$ as the projection matrix. From the point of view of subspace, the projection component can be viewed as a coefficient of subspace:
\begin{equation}
	\bs{C} = (\bs{V}^\intercal\bs{V})^{-1}\bs{V}^\intercal\Xa
\end{equation}
where $\mathbf{C}$ is the coefficient matrix.

Now, we need to use the subspace to reconstruct the features without the noise:
\begin{equation}
    \bs{Y} = \bs{U}\bs{C}
\end{equation}
where $\bs{Y}$ has the same size with $\Xa$. In practice, the output of the kernel function $f_{\theta}$ is $\bs{U}$. This reduces the amount of memory and computation when the input is large. 
}

\subsection{NBNet Architecture and Loss Function}

The architecture of NBNet is illustrated in Fig.~\ref{fig:architecture}(a). The overall structure
is based on a typical UNet~\cite{ronneberger2015u} architecture. NBNet has 4 encoder stages and 4 corresponding decoder stages, where feature maps
are downsampled to $\frac{1}{2}\times$ scale with a $4\times 4$-stride-$2$ convolution at the end of each encoder stage, and upsampled to $2\times$ scale with a $2\times 2$ deconvolution before each decoder stage. Skip connections pass large-scale low-level feature maps from each encoder stage to
its corresponding decoder stage. The basic convolution building blocks in encoder, decoder and skip connections follow the same residual-convolution structure depicted in Fig.~\ref{fig:architecture}(b).
We use LeakyReLU as activation functions for each convolutional layer.

The proposed SSA modules are placed in each skip-connection. As feature maps from low levels contain
more detailed raw image information, we take the low-level feature maps as $\Xa$ and high-level 
features as $\Xb$ and feed them into an SSA module. In other words, low-level feature maps from
skip-connections are projected into the signal subspace guided by the upsampled high-level features. The projected features are then fused with the original high-level feature before outputing to the next decoder stage.

Compared with conventional UNet-like architectures, which directly fuse low-level and 
high-level feature maps in each decoder stage, the major difference in NBNet is low-level features
are projected by SSA modules before fusion.

Finally, the output of the last decoder pass a linear $3\times 3$ convolutional layer as the global
residual to the noisy input and outputs the denoising result.

The network is trained with pairs of clean and noisy images, and we use simple $\ell_1$ distance 
between clean images and the denoising result as the loss function, written as:
\begin{equation}
    \mathcal{L}(G, \bs{x}, \bs{y}) = \|\bs{x}-G(\bs{y})\|_1 \text{,}
\end{equation}
where $\bs{x}$, $\bs{y}$ and $G(\cdot)$ represent clean image, noisy image and NBNet, respectively.

\remove{
In order to incorporate SSA with deep neural networks, we further propose the U-Net-based network structure SSA-U-Net and apply it to image denoising task. 

Firstly, we build a sub-network with two $3\times3$ convolutional layer and a $1\times1$ convolutional skip-connection for the SSA module. It provides stable training results and is a good choice for the kernel function $f_{\theta}$.

Then we put a SSA module in front of each up-sampled convolutional layer in the standard U-Net, and the SSA module takes as input features from two different layers, one is the skip-connection and the other is the up-sampled layer, see fig. Information extracted from coarse scale, the skip-connection features, usually contains more noise than that from fine scale, it needs to be transformed as $\Xa$. Then we take the information extracted from fine scale, the up-sampled layer features,  as $\Xb$.

Finally, in view of deep features are more useful for distinguishing signal from noise, we add convolutional layers at each skip-connection. More coarse scale the skip-connection is from the more convolutions are added.

For brevity, we define the entire network as $G$. Given a noisy image $y$ and a clean image $x$, the network outputs $G(y)$. We optimize the proposed network using the Charbonnier loss:
\begin{equation}
\mathcal{L}\left(x, y, G(y)\right)=\sqrt{\left\|x-(y - G(y))\right\|^{2}+\varepsilon^{2}}
\end{equation}
where $\varepsilon$ is a constant which we empirically set to $10^{-3}$ for all the experiments.
}


\section{Evaluation and Experiments}

\begin{table*}
    \centering
    \setlength{\tabcolsep}{2.5pt}
    \begin{tabular}{c c c c c c c c c c c c c}
        \toprule
        \multirow{3}*{Cases}&\multirow{3}*{Datasets} & \multicolumn{11}{c}{Methods} \\
        \cline{3-13}
        & & CBM3D & WNNM    & NCSR  &MLP  &DnCNN-B &MemNet  &FFDNet
        &$\text{FFDNet}_v$ &UDNet &VDN & Ours\\
        & & ~\cite{guo2018toward} & ~\cite{gu2014weighted}    & ~\cite{dong2013nonlocally}  &~\cite{burger2012image}  &~\cite{zhang2017beyond} &~\cite{tai2017memnet}  &~\cite{zhang2018ffdnet}
        &~\cite{zhang2018ffdnet} &~\cite{lefkimmiatis2018universal} &~\cite{yue2019variational} & \\
        
        \hline                                                        
        \multirow{3}*{$\sigma=15$} & Set5                                     
                     &33.42  &32.92   &32.57 &-     &34.04           &34.18       &34.30             &34.31    &34.19 &\textit{34.34} &  \textbf{34.64}\\
        \cline{2-13}                                                 
           &  LIVE1  &32.85  &31.70   &31.46 &-     & 33.72          &33.84       &\textit{33.96}    &\textit{33.96}    &33.74 &33.94 & \textbf{34.25}\\
        \cline{2-13}                                                 
           &  BSD68  &32.67  &31.27   &30.84 &-     &33.87  &33.76       &33.85             &33.68             &33.76 &\textit{33.90} & \textbf{34.15}\\
        \hline                                                        
        \multirow{3}*{$\sigma=25$} & Set5                                     
                     &30.92  &30.61   &30.33 &30.55 &31.88           &31.98       &32.10    &32.09             &31.82 &\textit{32.24} & \textbf{32.51} \\
        \cline{2-13}                                                  
           &  LIVE1  &30.05  &29.15   &29.05 &29.16 &31.23           &31.26       &31.37    &31.37    &31.09 &\textit{31.50} & \textbf{31.73}\\
        \cline{2-13}                                                 
           &  BSD68  &29.83  &28.62   &28.35 &28.93 &31.22  &31.17       &31.21             &31.20             &31.02 &\textit{31.35}& \textbf{31.54}\\
        \hline                                                       
        \multirow{3}*{$\sigma=50$} & Set5                                     
                     &28.16  &27.58   &27.20 &27.59 &28.95           &29.10       &29.25    &29.25    &28.87 &\textit{29.47} & \textbf{29.70}\\
        \cline{2-13}                                                 
           &  LIVE1  &26.98  &26.07   &26.06 &26.12 &27.95           &27.99       &28.10    &28.10    &27.82 &\textit{28.36}  & \textbf{28.55} \\
        \cline{2-13}                                                 
           &  BSD68  &26.81  &25.86   &25.75 &26.01 &27.91           &27.91       &27.95    &27.95    &27.76 &\textit{28.19}  & \textbf{28.35} \\
        \bottomrule
    \end{tabular}
    \vspace{-2.5mm}
    \caption{\small{The PSNR(dB) results of all competing methods on AWGN noise cases of three
test datasets.}}
    \label{tab:psnr_iidgauss}
\end{table*}

We evaluate the performance of our method on synthetic and real datasets and compare it with previous methods. Next, we describe the implementation details. Then we report results on five real image datasets. Finally, we perform ablation studies to verify the superiority of the proposed method.

\subsection{Training Settings}
The proposed architecture requires no pre-training and it can be trained through an end-to-end strategy. The number of subspace $K$ is set by experience to $16$ for all modules.

In the training stage, the weights of the whole network are initialized according to~\cite{he2015delving}. We use Adam~\cite{kingma2014adam} optimizer with momentum terms ($0.9$, $0.999$). The initial learning rate is set to $2\times 10^{-4}$ and the strategy of decreasing the learning rate is cosine annealing. The training process takes $700,000$ minibatch iterations.


During training, $128\times 128$-sized patches are cropped from each training pair as an instance, and
$32$ instances stack a mini-batch. We apply random rotation, cropping and flipping to the images to augment the training data.

\subsection{Results on Synthetic Gaussian Noise}
We first evaluate our approach on synthetic noisy dataset. We follow the experiment scheme
described in VDN~\cite{yue2019variational}. 
The training dataset includes 432 images from BSD~\cite{arbelaez2010contour}, 400 images from the validation set of ImageNet~\cite{deng2014scalable} and 4744 images from the Waterloo Exploration Database~\cite{ma2016waterloo}.
The evaluation test dataset are generated from Set5~\cite{kim2016accurate}, LIVE1~\cite{kim2016accurate} and BSD68~\cite{amfm_pami2011}.

In order to achieve a fair comparison, we use the same noise generation algorithm as~\cite{yue2019variational}, where non-i.i.d. Gaussian noise is generated by:
\begin{equation}
    \label{eqn:noniid}
    \bs{n}=\bs{n}^{1} \odot \bs{M}, \quad n_{i j}^{1} \sim \mathcal{N}(0,1),
\end{equation}
where $\boldsymbol{M}$ is a spatially variant mask. Four types of masks are generated, one for 
training and three for testing. By this way, the generalization ability of the noise reduction
model can be well tested.

Table~\ref{tab:psnr_noniid} lists the PSNR performance results of different methods on 
non-i.i.d Gaussian noise, where our NBNet method outperform the baseline VDN method on
every test case, although VDN has an automatic noise level prediction while our method
is purely blind noise reduction. More results on additive Gaussian white noise~(AWGN)
with various noise levels ($\sigma=15,25,50$) also indicates our method surpasses VDN
by an average margin of $\sim 0.3$ dB in PSNR.

Our noise reduction method does not explicitly rely on a prior distribution of noise
data, but it still achieve the best results in our evaluation. This shows the effectiveness
of the proposed projection method which helps separating signal and noise in feature space
by utilizing image prior.

\remove{
We evaluate our approach on synthetic noise data in this section. Similar to [\cite{YueYZM019}], we used training data included 432 images from BSD [\cite{}], 400 images from the validation set of ImageNet [] and 4744 images from The Waterloo
Exploration Database [] to train our network. For evaluation, Set5, LIVE1 and BSD68 in [] were adopted as test data. Following [], we generate the non-i.i.d. Gaussian noise as:
\begin{equation}
    \label{eqn:noniid}
    \boldsymbol{n}=\boldsymbol{n}^{1} \odot \boldsymbol{M}, \quad n_{i j}^{1} \sim \mathcal{N}(0,1),
    \end{equation}
where $\boldsymbol{M}$ is a spatial variant mask. Four types of masks are generated, one for training and three for testing. Based on this noise synthesis approach, the generalization ability of the model is widely explored.

VDN trains a network by generating noisy images following equation~\eqref{eqn:noniid} via assuming that training data are under a non. i.i.d. Gaussian distribution. Then VDN is tested on both i.i.d. and non-i.i.d. datasets separately. 

Folllowing VDN, for non-i.i.d testing data, Table~\ref{tab:psnr_noniid} illustrates the results. We then evaluate the performance of our method on additive Gaussian white noise (i.i.d.) with various levels ($\sigma$  = 15, 25, 50) of AWGN. Table~\ref{tab:psnr_iidgauss} lists the average PSNR results. Our model does not rely on a prior distribution of training data, however, better results are achieved compared to VDN. 

Our method obtains the best performance, demonstrating the ability to handle many different types of noise.
}

\subsection{Results on SIDD Benchmark}
\label{sec4.3}

The Smartphone Image Denoising Dataset (SIDD)~\cite{abdelhamed2018high} , are about 30,000 noisy images from 10 scenes under different lighting conditions using five representative smartphone cameras and generated their ground truth images through a systematic procedure. SIDD can be used to benchmark denoising performance for smartphone cameras. As a benchmark, SIDD splits 1,280 color images for the validation.

In this section, we use SIDD benchmark~\cite{abdelhamed2018high} to verify the performance of our method on a real-world noise reduction task. We compare with the previous methods, including VDN, DANet, and MIRNet. Table~\ref{table:sidd} illustrates a quantitative comparison between previous methods and ours in Fig~\ref{fig:sidd example}. We also provide visualization of noise reduction results from different models. The number of parameters and computational cost of each model are shown in Fig~\ref{fig:flop_psnr}. 

Compared to MIRNet, we provide 39.75 PSNR compared to MIRNet's 39.72 by only taking \textbf{11.2\%} of its computational cost and \textbf{41.82\%} of its number of parameters. In the SSIM metric, we have a performant rise over MIRNet, boosting from 0.959 to ours 0.969. This growth explains that our model concentrates further on regional textures and local features.

\begin{figure*}[h]
  \centering 
  \scalebox{1.0}{\hspace{-3mm}
    \begin{tabular}[t]{c@{ }c@{ }c@{ }c@{ }c@{ }}
    \includegraphics[width=.2\textwidth]{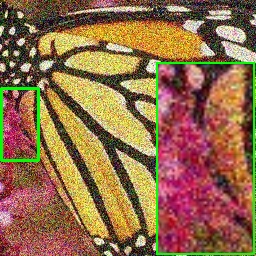}& \hspace{-1.4mm}
    \includegraphics[width=.2\textwidth]{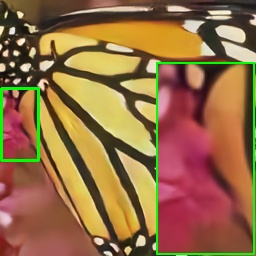}&  \hspace{-1.4mm}
    \includegraphics[width=.2\textwidth]{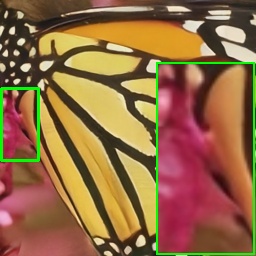}& \hspace{-1.4mm}
    \includegraphics[width=.2\textwidth]{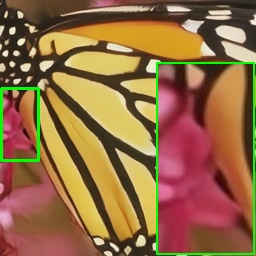}& \hspace{-1.4mm}
    \includegraphics[width=.2\textwidth]{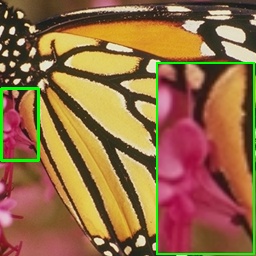}
    \vspace{0.2mm}
    \\
    \vspace{0.5mm}
     Noisy & FFDNet~\cite{zhang2018ffdnet} & VDN~\cite{yue2019variational} & Ours & Reference \\
    \end{tabular}
   }
    \vspace{-1mm}
    \caption{\small Results of Gaussian noise reduction. Our method obtains better visual results in the flower area.}
    \label{fig:sidd example}
 \end{figure*}

\subsection{Results on DND Benchmark}

The Darmstadt Noise Dataset (DND)~\cite{plotz2017benchmarking} consists of 50 pairs of real noisy images and corresponding ground truth images that were captured with consumer-grade cameras of differing sensor sizes. For every pair, a source image is taken with the base ISO level while the noisy image is taken with higher ISO and appropriately adjusted exposure time. The reference image undergoes careful post-processing involving small camera shift adjustment, linear intensity scaling, and removal of low-frequency bias. The post-processed image serves as ground truth for the DND benchmark.

We evaluate the performance of our method on the DND dataset which contains 50 images for testing. It provides bounding boxes for extracting 20 patches from each image, resulting in 1000 patches in total. Note that the DND dataset does not provide any training data, so we employ a training strategy by combining the dataset of SIDD and Renoir~\cite{anaya2014renoir}.  Results are submitted to the DND benchmark by utilizing the same model that provides the best validation performance on the SIDD benchmark.


Follow to MIRNet, we only use SIDD training set and the same augmentation strategy to train our NBNet. Table~\ref{table:dnd-result} shows the results of various methods, we can notice that NBNet can provide a better PSNR compared to MIRNet's $39.88$ dB with just a fractional of both computational cost and the number of parameters of MIRNet mentioned in section~\ref{sec4.3}. Visual results compared to other methods on DND are also provided in Fig~\ref{fig:dnd example}. Our method can provide a clean output image while preserving the textures and sharpness.


\subsection{Ablation Study}

\begin{table}[t]
    \centering
    \begin{tabular}{cccc}
    \toprule
    Method & \# Params & Comp. Cost &PSNR \\
         & ($\times 10^6$) & (GFlops) & (dB)\\
    \midrule
    UNet & 9.5 & 3.88 & 39.62\\
    UNet+SSA & 9.68 & 4.28 & 39.68\\
    UNet+Blocks & 13.13 & 21.8 & 39.69\\
    UNet+Blocks+SSA & 13.31 & 22.2 & 39.75 \\
    \bottomrule
    \end{tabular}
    \caption{Ablation study on SSA and other modules}
    \label{tab:abl study}
\end{table}
We examine three major determinants of our model: a) SSA module for another network, and b) the dimension of the signal subspace, i.e. the number of basis vectors $K$. c) the options about projection.
\subsubsection{Integrated into DnCNN}
For evaluating the effectiveness of our proposed SSA module, we consider another classical architecture DnCNN as a baseline. In order to use $\Xa$ and $\Xb$ shown in Equation \ref{eq: basis_func},  we regarded the feature of the first convolution as $\Xa$ and the feature before the last convolution as $\Xb$.  The results are shown in Table \ref{tab:abl dncnn}.  DnCNN + Concat achieves about 0.2dB higher than DnCNN by simply concatenating $\Xa$  and $\Xb$ to utilizing the different level features,  while the DnCNN + SSA  with our SSA module achieves about 0.5dB higher than DnCNN.

\begin{table}[t]
   \centering

   \begin{tabular}{c c c c c}
   \toprule
      K & K=1 & K=8 & K=16 & K=32 \\
   \midrule
   PSNR &  39.28 & 39.74 & 39.75 & -\\
   \bottomrule
   \end{tabular}
      \caption{Effects of subspace dimensionality K on SIDD. Our model does not converge when K=32}
    \label{tab:k exps}
\end{table}
\subsubsection{Influence about Different k Values}
Table~\ref{tab:k exps} provides the results on SIDD with different $K$ values. 
When the number of basis vectors $K$ is set to 32, our model does not converge. In this 
setting, as the number of channels in the first stage is also 32, the SSA module does
cannot work effectively as subspace projection since $K$ equals to the full dimension size. On the other hand, the higher dimension of the subspace may increase the difficulty of
model fitting, hence cause instability of training.
The rest experiments shows that the best choice of $K$ is 16. If $K$ equals 1, the information kept in the subspace is insufficient and cause significant 
information loss in the skip-connection. 
Setting $K$ to 8 and 16 leads to comparable performance, and the SSA module might create a low-dimensional, compact, or classifiable subspace. Therefore, we can see that the subspace dimension $K$ is a robust hyper-parameter in a reasonable range.

\subsubsection{Options about Projection}

In Table~\ref{tab:abl study proj}, we evaluate different options about projection: how to generate basis vectors and how to select feature maps for projection. Let's denote $Proj(a,b)$ as a projection operation where $a$ is projected to the basis generated based on $b$. As shown in first and second rows in Table~\ref{tab:abl study proj}, basis generation based  only on $\Xa$ makes training unstable, resulting in non-convergence. On the contrary, compare third and forth rows, basis generation based on only $\Xb$ enables the network to be trainable, but get unsatisfactory results. The best results are shown in the last two rows. The network achieves better performance by considering both $\Xa$ and $\Xb$. Therefore, projecting $\Xa$ on the basis generated by $\Xa$ and $\Xb$ obtains the best PSNR, which is 39.75 dB.
\begin{table}[t]
    \centering
    \begin{tabular}{ccc}
    \toprule
     & Method  & PSNR(dB)\\
    \midrule
    1 & DnCNN & 38.04 \\
    2 & DnCNN + Concat & 38.21\\
    3 & DnCNN + SSA & 38.59 \\
    \bottomrule
    \end{tabular}
    \caption{Ablation study on DnCNN architecture}
    \label{tab:abl dncnn}
\end{table}
\begin{table}[t]
    \centering
    \begin{tabular}{ccc}
    \toprule
     & Method  & PSNR(dB) \\
    \midrule
    1 & $Proj$($\Xa, \Xa$) & - \\
    2 & $Proj$($\Xa, \Xb$) & 39.02\\
    3 & $Proj$($\Xb, \Xb$) & 38.48\\
    4 & $Proj$($\Xb, \Xa$) & - \\
    5 & $Proj$($\Xb, \Xa \& \Xb$) & 39.68 \\
    6 & $Proj$($\Xa, \Xa \& \Xb$) & 39.75 \\
    \bottomrule
    \end{tabular}
    \caption{Ablation study on projections and '-' denotes non-convergence}
    \label{tab:abl study proj}
\end{table}

\subsection{Basis Visualization and Discussion}

\begin{figure}[t]
   \begin{center}
   \scalebox{0.95}{\hspace{-3mm}
   \begin{tabular}[b]{c@{ } c@{ }  c@{ } c@{ } c@{ }	}
        \multirow{4}{*}{\includegraphics[width=.33\textwidth,valign=tl]{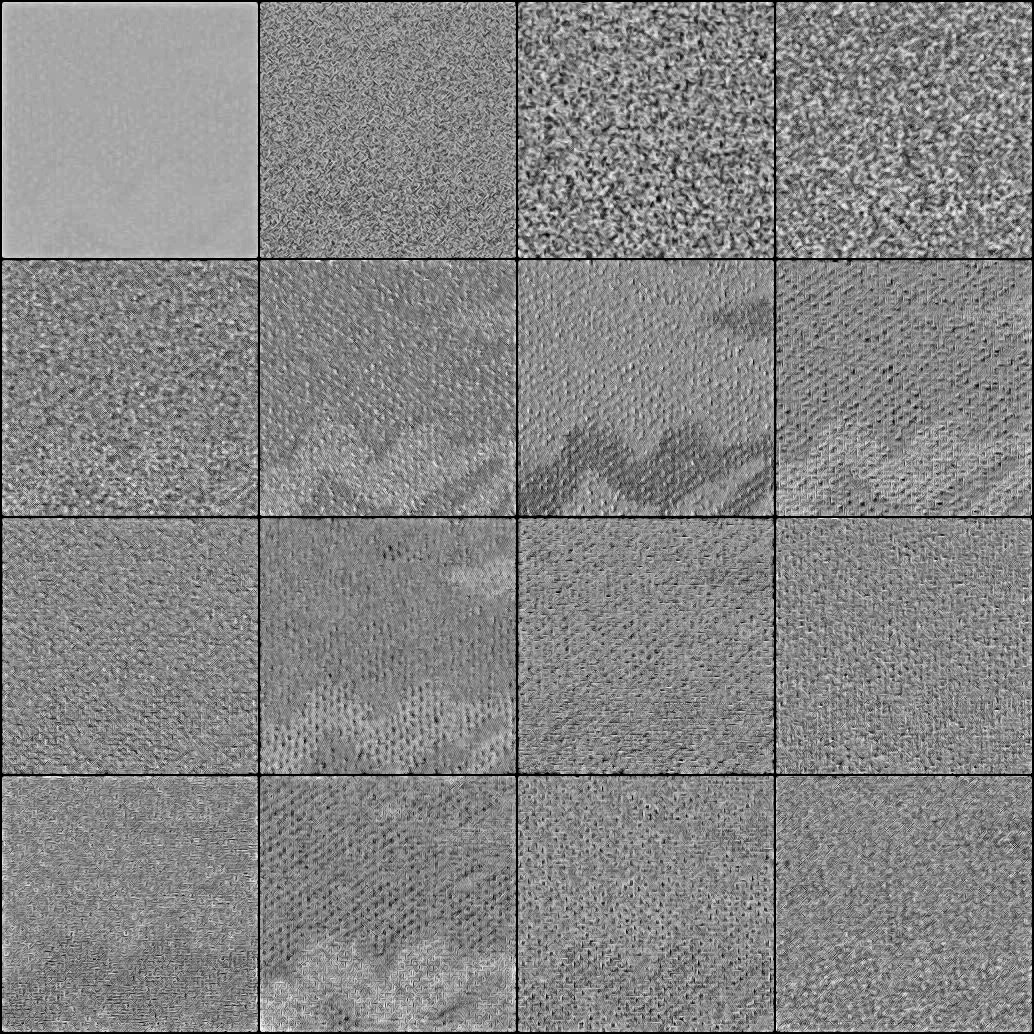}} 
       &  \includegraphics[width=.152\textwidth,valign=t]{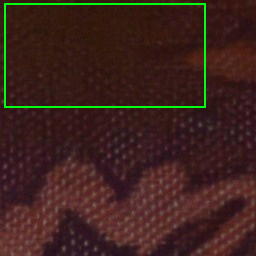}

       \vspace{0.5mm}
   \\
       &  UNet (31.81 dB)  \\

       &\includegraphics[width=.152\textwidth,valign=t]{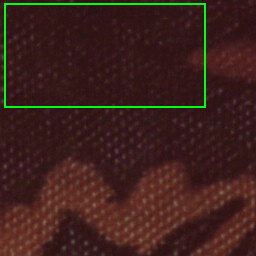}  
       
        \\
   
         Visualization of basis & SSA (32.15 dB)
        \\
              
   \end{tabular}}
   \end{center}
   \caption{\small {\textbf{Left}: the basis vectors that span the projection subspace. It can be seen that the dotted pattern is captured in the channels. \textbf{Right}: denoising results with and without the SSA module. When SSA is used, the weak texture in the upper part is recovered better and appear more consistent with other parts of the image.
   } }
   \label{fig:basis vector visual}
   
\end{figure}

To gain insight about how the learned subspace projection works , we pick a sample image and inspect the subspace generated by the SSA module. Fig~\ref{fig:basis vector visual} plots the 16 basis vectors together with the prediction with and without the SSA module. 
It can be seen that when SSA is enabled, the dotted texture in the dark region is recovered in a way consistent with other part of the patch. This is different when SSA is disabled: the network simply blurs the upper area. Same phenomenon is also observed
in Fig~\ref{fig:sidd example} where NBNet outperforms other methods in weak-textured regions.

Not surprisingly, this phenomenon finds its root in the projection basis vectors. 
As shown in the left side of Fig~\ref{fig:basis vector visual}, many of the 16 channels contain the dots  pattern that evenly span the whole image patch. 
We can thus reasonably surmise that this 
improvement should be attributed to the non-local correlation created by the 
SSA module: the weak textures on the upper part are supported by 
the similar occurrence in other parts of the image, and the projection reconstructs
the texture by combining the basis with globally determined coefficients.
A conventional convolutional neural network, on the contrary, rely on responses
of fixed-valued local filters and coarse information from downsampled features. 
When the filter response is insignificant and coarse information is blurry, 
e.g. in the weak texture areas, non-local information can hardly improve local responses.  

\section{Conclusion}
In this study, we revisit the problem of image  denoising and provide a new prospective of subpsace projection. Instead of relying on complicate network architecture or accurate image noise modeling, the proposed subspace basis generation and projection operation can naturally introduce global structure information into denoising process and achieve better local detail preserving. We further demonstrate such basis generation and projection can be learned with SSA in end-to-end fashion and yield better efficiency than adding convolutional blocks. We believe subspace learning is a promising direction for image denoising as well as other low-level vision tasks, and it is worth further exploration. 

\section*{Acknowledgement}
This research was supported in part by National Natural Science Foundation of China (NSFC) under grants No.61872067 and No.61720106004, in part by Research Programs of Science and Technology in Sichuan Province under grant No.2019YFH0016.

\clearpage

\bibliographystyle{ieee_fullname}
\bibliography{egbib}

\end{document}